\def\CatchManArxivVersion{1}
\RequirePackage[T1]{fontenc}
\documentclass[letterpaper, 10 pt, conference]{ieeeconf}

\usepackage{mathptmx}
\usepackage{graphicx}
\usepackage{amsmath}
\usepackage{amssymb}
\usepackage{booktabs}
\usepackage{array}
\usepackage{xcolor}
\usepackage{placeins}
\usepackage[colorlinks=true, allcolors=blue]{hyperref}
\usepackage{cleveref}
\usepackage{xspace}

\ifdefined\CatchManArxivVersion
  \hypersetup{
    pdfauthor={Taoyang Jia, Weikai Huang, Linxin Song, Jared Darlington, Jieyu Zhao, Yue Wang, Jiafei Duan, Zhongzheng Ren, Ranjay Krishna},
    pdfsubject={Real-time humanoid box catching with egocentric vision},
    pdfkeywords={humanoid robotics, egocentric vision, reactive control, box catching}
  }
\else
  \hypersetup{pdfauthor={}, pdfsubject={}, pdfkeywords={}}
\fi

\definecolor{wkblue}{RGB}{0,90,180}
\newif\ifwkmarks
\wkmarkstrue
\DeclareRobustCommand{\wk}[1]{\ifwkmarks{\color{wkblue}\textbf{[WK:} #1\textbf{]}}\fi}

\definecolor{paperpink}{RGB}{174,46,105}
\definecolor{papergray}{RGB}{96,96,96}
\newlength{\paperintegerwidth}
\newlength{\paperfractionwidth}
\newlength{\papersdintegerwidth}
\newcommand{\papertablestyle}{%
  \fontsize{7.5}{9}\selectfont
  \setlength{\tabcolsep}{2.2pt}%
  \renewcommand{\arraystretch}{1.15}%
  \setlength{\heavyrulewidth}{0.7pt}%
  \setlength{\lightrulewidth}{0.4pt}%
  \setlength{\cmidrulewidth}{0.4pt}%
  \setlength{\aboverulesep}{0.4ex}%
  \setlength{\belowrulesep}{0.6ex}%
  \settowidth{\paperintegerwidth}{\textbf{100}}%
  \settowidth{\paperfractionwidth}{\textbf{0}}%
  \settowidth{\papersdintegerwidth}{9}%
}
\newcommand{\tableours}[1]{\textcolor{paperpink}{#1}}
\newcommand{\tableref}[1]{\textcolor{papergray}{#1}}
\def\paperdecimal#1.#2\relax{%
  \makebox[\paperintegerwidth][r]{#1}.\makebox[\paperfractionwidth][l]{#2}}
\def\papersddecimal#1.#2\relax{%
  \makebox[\papersdintegerwidth][r]{#1}.\makebox[\paperfractionwidth][l]{#2}}
\newcommand{\tablenum}[1]{\mbox{\paperdecimal#1\relax}}

\newcommand{\tablebestnum}[1]{{\bfseries\tablenum{#1}}}

\newcommand{\papertablenote}[1]{%
  \par\vspace{3pt}%
  \begin{minipage}{\linewidth}\fontsize{7}{8.4}\selectfont\raggedright
  #1\end{minipage}%
}

\ifdefined\CatchManArxivVersion
  \wkmarksfalse % Hide review comments in the public version.
\fi

\newcommand{\model}[0]{\texttt{Reflex}\xspace}

\ifdefined\CatchManArxivVersion
\title{\raisebox{-0.3\height}{\includegraphics[height=2.4em]{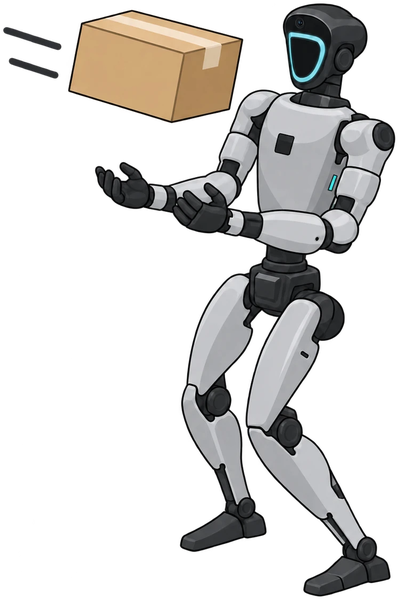}}\hspace{0.3em}\model: Visually Grounded Reactive Humanoid Control}
\else
\title{\model: Visually Grounded Reactive Humanoid Control}
\fi

\ifdefined\CatchManArxivVersion
\author{Taoyang Jia$^{*1,2}$, Weikai Huang$^{*1}$, Linxin Song$^{*3}$, Jared Darlington$^{1}$,\\
Jieyu Zhao$^{3}$, Yue Wang$^{3}$, Jiafei Duan$^{4}$, Zhongzheng Ren$^{5,6}$, and Ranjay Krishna$^{1}$\\
{\normalsize $^{1}$University of Washington \quad $^{2}$Stanford University \quad $^{3}$University of Southern California}\\
{\normalsize $^{4}$National University of Singapore \quad $^{5}$University of North Carolina at Chapel Hill \quad $^{6}$Allen Institute for AI}%
\thanks{$^{*}$Equal contribution (co-first authors).}%
}

\else
  \author{\phantom{Anonymous Authors}}
\fi

\IEEEoverridecommandlockouts
\makeatletter
\IEEEaftertitletext{%
  \begin{minipage}{\textwidth}
    \centering
    \def\@captype{figure}% Use the class's normal caption style without a float.
    \includegraphics[width=\textwidth]{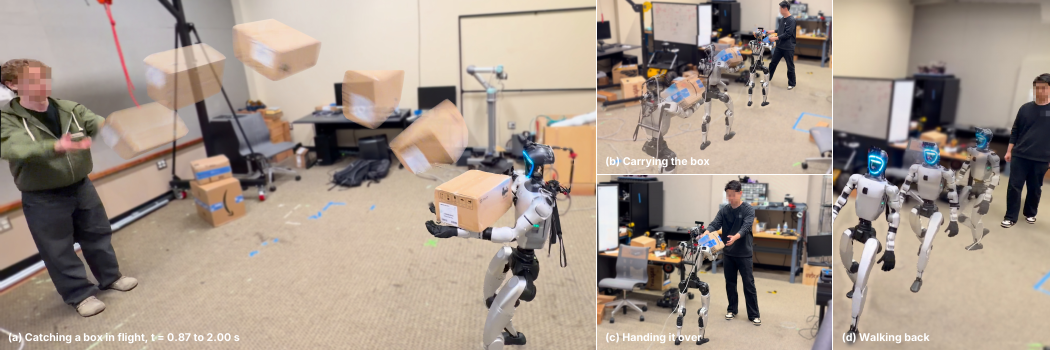}
    \caption{\textbf{Visual reactive whole-body catching on a humanoid.} (a) Catching a thrown carton from onboard RGB-D input. (b)--(d) Downstream application: carrying the box 7.8\,m, handing it to a person, and walking 7.3\,m back. The retained box is passed directly to the downstream walking controller for carrying and handover.%
    }
    \label{fig:teaser}
  \end{minipage}%
  \vspace{4pt}%
}
\makeatother

\begin{document}

\maketitle
\thispagestyle{empty}
\pagestyle{empty}

\begin{abstract}
Humans effortlessly react to the world; they adeptly return a tennis shot, catch a falling cup, step aside from a passing cart, or pass a box tossed during truck unloading. These everyday actions require anticipating a moving target and reflexively responding with the whole body within about a second. Such visual reactive whole-body control remains challenging for humanoids, which must (1) coordinate whole-body control, (2) react to object dynamics in real time, and (3) ground perception in egocentric vision. 
We study catching thrown boxes as a representative task that demands all three capabilities; existing systems cover only part of it, relying on end-effectors, motion capture, or marked objects.
We present \model, a three-stage framework that learns these capabilities stage by stage while connecting them through a compact representation of object dynamics. First, a state-based policy learns whole-body catching with reinforcement learning. Second, the policy learns to infer the dynamics of a moving box from delayed and incomplete observations. Third, an egocentric RGB-D encoder learns to recover this dynamics representation directly from visual observations while keeping the whole-body controller fixed. This decomposition provides each capability with a direct training signal and makes visual learning substantially more scalable.
In simulation, \model achieves 85.7\% catch success from egocentric RGB-D, compared with 87.9\% when given privileged box observations. Temporal visual history is critical: removing it reduces success by 35.6 and 39.8 points on centered and laterally offset throws. On a real Unitree G1, \model catches human-thrown boxes with 65\% and 40\% success at 2.5 and 5\,m, reaching 76\% and 59\% of human catchers' success at the same distances.
\end{abstract}
\section{Introduction}
\label{sec:introduction}

Humans routinely react to fast-moving objects with their entire body. 
A goalkeeper shifts, dives, and braces for an incoming shot; a worker catches a box tossed from a truck; a person reaches for a falling cup while stepping to keep balance; another twists aside as a cart passes unexpectedly close. 
These behaviors require more than tracking an object in motion. They require predicting where it will go, deciding how the body should meet or avoid it, and coordinating a whole-body response before contact, often within about a second. Humanoids have the morphology to perform such behaviors, but robust \emph{visually reactive whole-body control} remains largely out of reach.

This capability couples three challenges. First, the robot must control the whole body: it must place its feet and torso, move its arms into position, absorb impact, and remain balanced. Second, it must reflexively react to fast object dynamics under latency. In our real-robot trials, a thrown box reaches the robot roughly one second after release, while camera-to-control latency alone has a median of 69\,ms in our detector-based pipeline. The robot therefore cannot simply react to the latest observation; it must infer where the object is going and commit to an interception before contact. 
Third, it must do so from egocentric vision. Because the camera moves with the robot, observed image motion combines object motion with self-motion, while the robot's own arms increasingly occlude the object near contact.

We study catching thrown boxes as a representative test of all three challenges. Catching a box is more demanding than intercepting a ball with the hands: the robot must reposition its body, use the forearms and torso to form a support, absorb the box's momentum, and retain the load without losing balance. At the same time, the box is an ordinary uninstrumented object observed only through onboard RGB-D cameras. The task is difficult even for people: across our trials, three human catchers succeed on only 28.3\% of throws from 7\,m.

Existing systems address only parts of this problem. Robotic manipulators can catch moving objects from fixed or wheeled bases, avoiding the challenge of whole-body balance and impact absorption~\cite{catchingflight,agilecatching,catchit}. Recent humanoid systems demonstrate dynamic interactions such as punching away a ball~\cite{humanoidgoalkeeper} or catching a basketball with the hands~\cite{humanx}, but rely on motion capture or retroreflective objects rather than grounding fast object motion from onboard vision. 
As a result, whole-body control, fast prediction of object dynamics, and egocentric visual grounding have rarely been combined on a real humanoid.

\begin{figure*}[t]
\centering
\includegraphics[width=\textwidth]{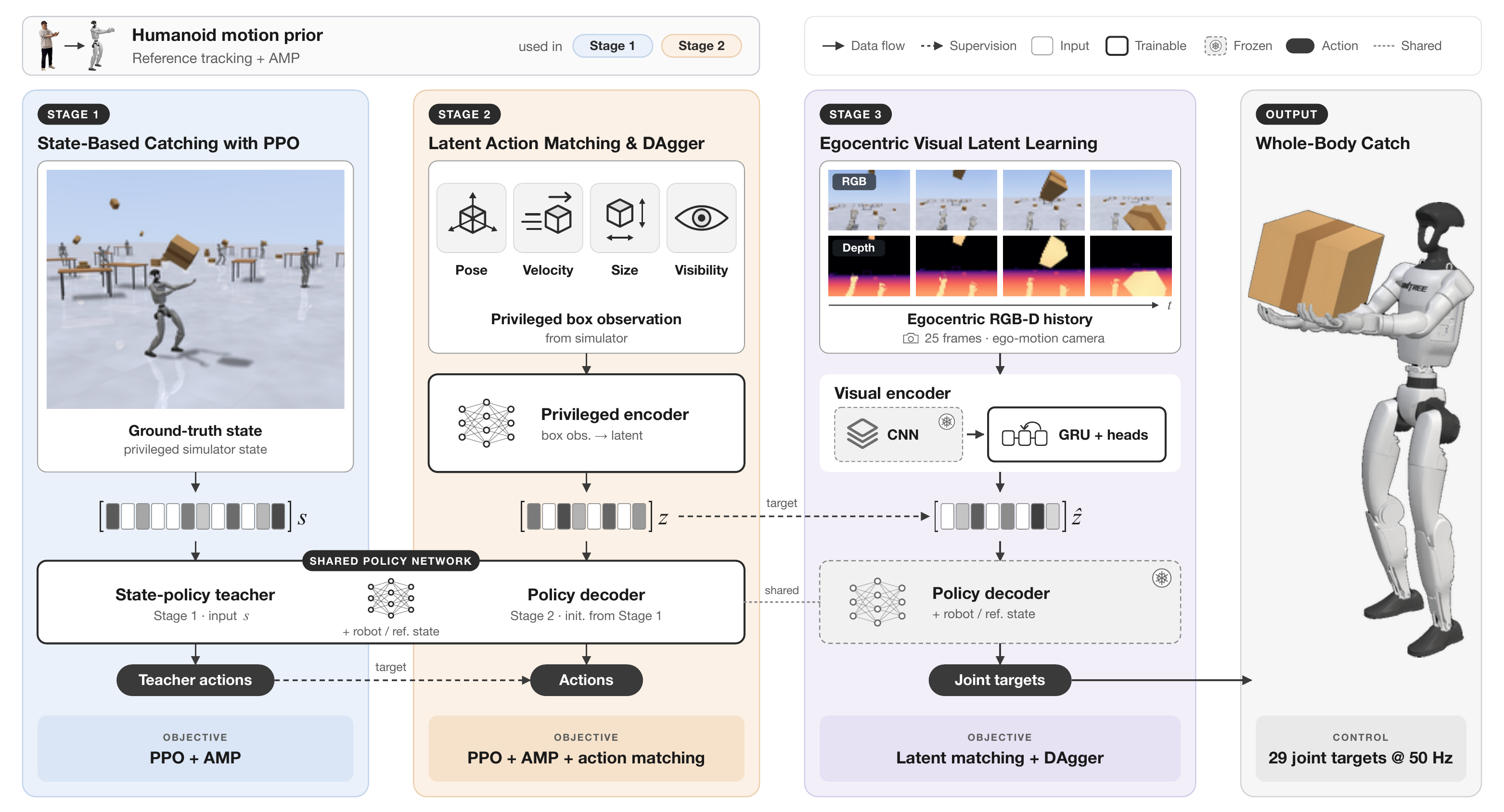}
\caption{\textbf{Training pipeline and policy architecture.} Stage~1 learns whole-body control by training the state-policy teacher with PPO. Stage~2 learns the object dynamics: it initializes the privileged encoder and decoder latent-input weights by matching teacher actions, then jointly refines the encoder and decoder with PPO. Human motion examples supply reference tracking and an AMP style reward. Stage~3 learns visual grounding by predicting the latent from RGB-D histories and camera ego-motion, using box-state supervision and DAgger rollouts. The pretrained CNN and heatmap head stay fixed during GRU training; the privileged encoder and decoder remain fixed throughout Stage~3. The decoder also receives robot/reference observations. Because Stage~3 learns only the latent predictor, the visual controller reuses the PPO-trained decoder without modification.}
\label{fig:framework}
\vspace{-2pt}
\end{figure*}

We present \model, a three-stage framework for visually reactive whole-body control (see Figure~\ref{fig:framework}). \textbf{Stage~1} learns whole-body catching from privileged simulator state using reinforcement learning and human motion priors. \textbf{Stage~2} learns to infer the dynamics of a moving box from delayed (20--60\,ms latency) and incomplete observations of its position, velocity, orientation, size, and observation age, producing a latent representation that conditions the controller to react. \textbf{Stage~3} learns to recover this same dynamics representation from egocentric RGB-D history and camera ego-motion while keeping the whole-body controller fixed. Our decomposition gives each capability a direct training signal and isolates expensive visual learning from reinforcement learning.

In simulation, \model achieves 85.7\% catch success from egocentric RGB-D, compared with 87.9\% when given privileged box observations. Temporal visual history is vital for good performance; removing it reduces success by 35.6 and 39.8 points on centered and 0.6\,m-offset throws. On a real Unitree G1, \model catches human-thrown boxes on 65\% and 40\% of throws at 2.5 and 5\,m, reaching 76\% and 59\% of the human success rate at the same distances (85.0\% and 68.3\%). Ablations further show that camera ego-motion matters for visual grounding, that regressing the learned latent works better than regressing the full privileged box state, and that action-matching initialization outperforms weight mapping by 5.2 points on average.

More broadly, our results suggest a path toward humanoids that can react to dynamic events directly from onboard vision, rather than relying on instrumented environments or task-specific perception systems. Catching is one instance of a broader class of visually reactive whole-body skills, including interception, avoidance, handoffs, and physical assistance. These everyday tasks require predicting how the world will change and coordinating the body before contact.

\section{Related work}
\label{sec:related}

We review prior work on catching thrown objects, on humanoid whole-body control and loco-manipulation, and on privileged learning for visual sim-to-real transfer.

\subsection{Catching thrown objects}
Catching a thrown object requires predicting its trajectory and moving into position within a short time window. Early systems addressed this with fixed-base arms guided by vision~\cite{hove}, with later work learning object dynamics and catching motions from demonstration~\cite{catchingflight} or planning impact-aware bimanual motions with fixed-base arms~\cite{bimanualcatch}. Learning-based approaches have since extended catching to whole-body model predictive control combined with policy learning~\cite{agilecatching}, dexterous and bimanual throw-and-catch~\cite{dexcatch,dynhandover}, and mobile manipulators~\cite{catchball,catchit}. More recently, bipedal humanoids have been trained to intercept balls as goalkeepers~\cite{humanoidgoalkeeper} and to catch basketball passes~\cite{humanx}. However, most of these systems observe the object with motion capture or fixed external cameras, with fewer relying on onboard sensing: humanoids have kicked and struck moving balls from onboard vision~\cite{reactivesoccer,smash}, a legged manipulator plays badminton with a learned perception-error model~\cite{leggedbadminton}, and onboard catching has been shown on a wheeled humanoid~\cite{justin} and a quadruped~\cite{eventcatch}, and concurrent work catches a thrown box on a humanoid from egocentric RGB without reporting real-robot success rates~\cite{workhorse}. To address this, we train a humanoid to catch and retain a thrown box using onboard egocentric RGB-D.

\subsection{Humanoid whole-body control and loco-manipulation}
Human motion data has become a common foundation for humanoid whole-body control. This approach began in physics-based character animation with motion imitation~\cite{deepmimic} and adversarial motion priors~\cite{amp}. Extending this to object interaction, Catch \& Carry~\cite{catchcarry} paired a motor-primitive module learned from human demonstrations with an egocentric-vision policy trained by reinforcement learning to catch balls and carry boxes in simulation. These methods have since moved from simulation to hardware, where learned motion trackers provide general whole-body control on real humanoids~\cite{h2o,beyondmimic,sonic}. Recent loco-manipulation systems go further, lifting and carrying boxes and bulky objects through whole-body contact~\cite{digitbox,omniretarget,hdmi,physhsi,resmimic,embracing}. In these tasks, though, the object is at rest when interaction begins, so the robot can approach and make contact at its own pace. Catching removes this freedom, since the body must already be in place when the box arrives. Like these methods, we build on human motion, but train a dedicated catching controller that reacts to the incoming box and coordinates interception and retention.

\subsection{Privileged Learning and Visual Sim-to-Real}
Policies trained in simulation can use information that is unavailable at deployment, such as the exact state of an object. Privileged learning exploits this by training a teacher with such information to supervise a student that uses onboard sensors~\cite{lbc}, typically alongside asymmetric critics~\cite{asymmetric}, domain randomization~\cite{domainrand}, and on-policy data aggregation~\cite{dagger}. Legged robots adopted this approach early, with students that reproduce a teacher's latent from observation history~\cite{quadterrain,rma} or walk from egocentric depth~\cite{egolocomotion}. Humanoids now use it to perform loco-manipulation, door opening, parkour, and ladder climbing from depth or RGB~\cite{viral,doorman,parkour,ladderman}. However, these students perceive static terrain or slow-moving objects. A thrown box instead moves quickly through the view of a moving camera. Following RMA~\cite{rma}, we keep the action decoder fixed and train a visual model by latent regression to predict the privileged latent from RGB-D history and camera ego-motion.

\section{Task Formulation}
\label{sec:system}
We first define the catching task and its sensing and control interface, then the criteria used to evaluate it.

\subsection{Visually Grounded Reactive Catching}

We study the task of catching a human-thrown box with a 29-degree-of-freedom (DoF) Unitree G1. The robot must move into the path of the incoming box, receive it on its forearms against the torso, dissipate its momentum, and retain it without falling. Unlike quasi-static loco-manipulation, the robot cannot choose when contact occurs: it must infer the box trajectory and coordinate its whole body before the box arrives.

At deployment, the visual input is a stream of $192\times144$ RGB-D observations $\mathbf I_t$ from a torso-mounted camera. The controller additionally receives the robot state required for whole-body stabilization, but no box pose, velocity, or external-camera measurement. The camera moves with the robot and the box becomes increasingly occluded near contact, making the task a closed-loop coupling of visual estimation and whole-body response.

At each 20\,ms control step, the model outputs a 29-dimensional whole-body joint command, corresponding to one command for each actuated DoF. These commands are executed as joint-position targets. Thus, the task-level interface maps egocentric RGB-D observations and robot state directly to 29-DoF control; the internal policy decomposition is introduced only in Sec.~\ref{sec:method}. Simulation uses Isaac Lab~\cite{isaaclab}, with physics at 200\,Hz and control at 50\,Hz.

\subsection{Success Criteria}
\label{sec:success-metrics}

We evaluate three distinct outcomes. \emph{Interception} indicates that the robot has arrested the incoming motion of the box: the box slows below 1\,m/s near the torso, regardless of whether it is subsequently retained. The primary metric, \emph{catch success}, additionally requires the box to be supported in the chest region, with the wrists on opposite sides of the box, without relying on the head or shoulders, and at low speed for 0.6\,s. This criterion distinguishes sustained retention from incidental contact or a momentary trap.

We assess post-catch stability separately because retaining the box does not necessarily leave the robot in a settled posture. \emph{Torso excursion} records whether the torso tilt reaches $38^{\circ}$ between launch and the end of the episode, while falls are reported independently. For the box-size sweep, where a fixed chest-region test is not comparable across geometries, we use the geometry-agnostic sustained-hold criterion specified in Sec.~\ref{sec:robustness}. These metrics separate perception and interception failures from retention and whole-body stabilization failures.

\section{Learning Egocentric Whole-Body Catching}
\label{sec:method}

\paragraph{Perception--control decomposition}
We decompose the deployed policy into a visual perception function $V_\psi$ and a whole-body control function $\pi_\theta$. Given a history of $K$ egocentric RGB-D observations and camera ego-motion, $V_\psi$ predicts an eight-dimensional object-dynamics latent. The controller combines this latent with the robot/reference observation $\mathbf r_t\in\mathbb R^{160}$ and produces a normalized 29-DoF action:
\begin{equation}
\widehat{\mathbf z}_t
=V_\psi(\mathbf I_{t-K+1:t},\mathbf e_{t-K+1:t}),\qquad
\mathbf a_t=\pi_\theta(\mathbf r_t,\widehat{\mathbf z}_t).
\label{eq:visual}
\end{equation}
The action is converted to nominal joint-position targets as
\begin{equation}
\mathbf q_t^{\mathrm{nom}}=\mathbf q_0+\mathbf s\odot\mathbf a_t,
\label{eq:control}
\end{equation}
where $\mathbf q_0\in\mathbb R^{29}$ is the nominal joint configuration, $\mathbf s\in\mathbb R^{29}$ contains per-joint action scales, and $\odot$ denotes elementwise multiplication. This latent interface separates the question of what object information can be inferred from vision from how that information should drive the body.

Training $V_\psi$ and $\pi_\theta$ jointly from pixels with reinforcement learning is both computationally expensive and difficult to optimize. Every on-policy rollout would require high-throughput RGB-D rendering, while the policy must simultaneously explore a large visual space and a 29-dimensional action space. Catch success is delayed until contact and retention, providing weak supervision for learning object motion from images; camera ego-motion, intermittent occlusion, and policy-induced viewpoint changes further couple perception errors to the state distribution visited by the controller. A monolithic end-to-end policy would therefore entangle visual representation learning with motor-skill discovery and make perception adaptation require retraining the controller.

We instead organize learning into \textbf{three stages}, one per capability (Fig.~\ref{fig:framework}): \textbf{Stage~1} learns whole-body control from box state with PPO, \textbf{Stage~2} learns the object dynamics as a compact control latent (its PPO also refines the controller), and \textbf{Stage~3} grounds this latent in vision while the controller remains fixed. A downstream carrying and handover module follows the catching controller (Sec.~\ref{sec:deployment}).

\subsection{Stage 1: Learning Whole-Body Control}
\label{sec:state-policy}
Stage~1 isolates the motor skill: given box-state channels, PPO learns body placement, receiving posture, impact absorption, and retention in rendering-free simulation.

\paragraph{State observations and policy}
We train a state-based catching policy $\pi_{\mathrm{state}}$ with proximal policy optimization (PPO)~\cite{ppo} over successive warm-started runs. Its 170-dimensional observation $\mathbf o_t^{\mathrm{state}}$ combines the robot/reference input $\mathbf r_t$ with ten box-state channels: pelvis-frame position and linear velocity (three each), observation age (one), and indicators of measurement jitter, freshness, and frozen observations (three). Simulation supplies these channels by corrupting box detections with randomized noise, missed observations, and delay and passing them through the Kalman filter used on hardware (gravity and linear drag). The actor is a multilayer perceptron with hidden widths 512--256--128, exponential linear unit activations, and the 29 outputs of Eq.~\eqref{eq:control}; an asymmetric critic additionally observes uncorrupted full-body poses and ground-truth box position and velocity.

\paragraph{Catching objective and randomized training}
Approach shaping encourages interception, hold shaping rewards forearm support near the chest, and a sustained-catch bonus rewards retention. Tracking retargeted human reference motion and an adversarial motion prior (AMP)~\cite{amp} guide coordinated movement; penalties discourage excessive torso motion, abrupt commands, and falls. PPO updates the actor and critic, while the AMP discriminator learns from simulated motion and retargeted human examples. Throws, contact properties, and robot dynamics are randomized; held-box initializations and perturbations train retention and recovery.

\paragraph{Teacher for latent transfer}
After state-based PPO, we freeze the teacher. For Stage~2 action supervision, we query it with ground-truth position and velocity channels. Its decoder weights initialize the latent controller, and its actions label the initialization dataset. The ground-truth state-policy reference in the experiments instead feeds ideal (complete, noise-free, zero-delay) box measurements through the policy's Kalman filter.

\subsection{Stage 2: Learning Object Dynamics}
\label{sec:privileged-interface}
Stage~2 learns what the controller needs to know about the moving box: a privileged encoder compresses a delayed, partially observed box observation into a latent that PPO shapes to account for the box's motion, defining the object dynamics that Stage~3 must perceive.

\paragraph{Privileged latent interface}
The privileged encoder $E_\phi$ maps a box observation $\mathbf p_t\in\mathbb R^{26}$ to the latent used by the control function $\pi_\theta$:
\begin{equation}
\mathbf z_t=E_\phi(\mathbf p_t)\in[-1,1]^8.
\label{eq:latent}
\end{equation}
The 26 channels comprise camera-frame position (3), relative linear velocity (3), box half extents (3), orientation (12), angular velocity (3), visibility (1), and normalized observation age (1). Orientation uses the six unique entries of each axis outer product for the box's long and short axes, making the representation invariant to axis-sign flips.

We select the freshest visible sample in a delayed 25-frame window, with field-of-view masking and observation dropout. When no sample is visible, position, linear/angular velocity, and orientation terms are zeroed, visibility is zero, and normalized age is one; box half extents remain available. A linear projection and residual MLP, followed by $\tanh$, produce the bounded latent, whose coordinates are learned through action matching and task optimization. The encoder thus sees one delayed sample, including velocity, at a time; inferring motion from observation history is left to Stage~3.

\paragraph{Action-matching initialization}
Let $\mathbf a_t^{\mathrm T}=\pi_{\mathrm{state}}(\mathbf o_t^{\mathrm{state}})$ denote the teacher mean action. The action-supervision dataset $\mathcal D_A$ stores tuples $(\mathbf r_t,\mathbf p_t,\mathbf a_t^{\mathrm T})$.
We retain the teacher's decoder body, replace its object-input weights with randomly initialized latent-input weights, and initialize the new pathway by behavior cloning (BC):
\begin{equation}
\mathcal L_{\mathrm{BC}}=\mathbb E_{\mathcal D_A}\Big[
w_t^{\mathrm{vis}}\,\big\|\pi_\theta(\mathbf r_t,E_\phi(\mathbf p_t))
-\mathbf a_t^{\mathrm T}\big\|_2^2\Big].
\label{eq:init}
\end{equation}
The expectation is over dataset samples, and $w_t^{\mathrm{vis}}$ is a visibility-based sample weight: 5 for visible-box samples and 1 otherwise. Only the privileged encoder and the first decoder layer's latent-input weights are updated; the remaining decoder parameters stay fixed.

\paragraph{Action-labeled data aggregation}
We first collect teacher-driven trajectories, then run the latent controller and query the fixed state-policy teacher for action labels on its visited states. This dataset aggregation (DAgger)-style procedure~\cite{dagger} expands $\mathcal D_A$ with states encountered under latent control, and action matching is repeated on the aggregated data. These teacher action labels are distinct from the retargeted human motion examples used for reference tracking and the motion prior.

\paragraph{Task-specific PPO refinement}
After action matching and aggregation, PPO jointly optimizes the privileged encoder and the full decoder, using the catching objective and motion priors introduced in Stage~1. The asymmetric critic retains access to uncorrupted body poses and ground-truth box state while the actor receives $\mathbf r_t$ and the dynamics latent computed from delayed box observations. Teacher action matching is used for initialization and aggregation; PPO refinement optimizes task return and, because the decoder is also updated, refines whole-body control as well (Sec.~\ref{sec:exp-wbc}).

\begin{figure*}[t]
\centering
\includegraphics[width=\textwidth]{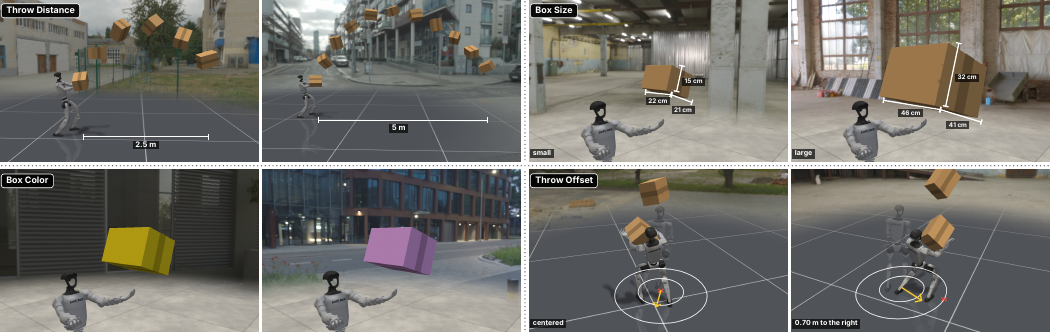}
\caption{\textbf{Throw and appearance variations.} Simulation examples show changes in release distance, box size, color, and landing location. The box-size panel illustrates measured real boxes; simulated weight and distance tests use the 34$\times$26$\times$22\,cm training box. Color augmentation is applied when training the visual component. These variations define the operating conditions evaluated in Tab.~\ref{tab:robustness}.%
% \textcolor{blue}{JF: for each example just show 2 will do, then it will look bigger too}
}
\label{fig:training}
\vspace{-2pt}
\end{figure*}

Latent-controller training extends the box-size distribution while retaining throw, dynamics, contact, and recovery randomization. Stage~2 samples object-observation latency once per episode from one to three control steps (20--60\,ms), training the controller to act on delayed perception; the camera-to-control latency of our detector-based hardware pipeline (69\,ms median; Tab.~\ref{tab:latency}) exceeds this range, and Stage~3 targets add a further 40\,ms lag. In a single-seed comparison, a controller trained without it is 5--11 points lower at every tested delay (Sec.~\ref{sec:exp-dynamics}). Figure~\ref{fig:training} illustrates the variations; numerical settings are given in Sec.~\ref{sec:training-settings}. The resulting encoder and decoder are frozen for Stage~3.

\subsection{Stage 3: Learning Visual Grounding}
\label{sec:visual-architecture}
Stage~3 grounds the dynamics latent in egocentric vision by supervised regression from RGB-D and ego-motion histories; rendering enters only this stage, and the controller remains unchanged.

\paragraph{Temporal visual architecture}
The visual encoder $V_\psi$ implements the perception function in Eq.~\eqref{eq:visual} using histories of length $K=25$. Visual control retains the robot/reference input, decoder, and joint-command interface; only the source of the dynamics latent changes.

A convolutional neural network (CNN) predicts a box-localization heatmap. A fixed geometric layer forms $\mathbf g_t\in\mathbb R^8$ from normalized image coordinates (2), local minimum depths in two neighborhoods (2), a heatmap peak score (1), and a camera-frame surface point obtained by pinhole backprojection (3). A gated recurrent unit (GRU) processes these features and nine-dimensional ego-motion vectors to infer object motion under viewpoint changes and missed detections. Appearance thus reaches the GRU only through the localization heatmap; in simulation, ego-motion comes from the simulator. Two output heads predict the dynamics latent and auxiliary position/velocity; the latter supplies training supervision, while only the latent drives the decoder.

\paragraph{Two-step visual training}
The Stage~2 privileged encoder $E_\phi$ and action decoder $\pi_\theta$ stay fixed throughout visual training. We train the visual encoder in two steps. First, we train the RGB-D CNN and box-localization heatmap using Gaussian targets at projected box centers with binary cross-entropy supervision. Second, we freeze the CNN and heatmap head and train the GRU to predict the dynamics latent and auxiliary box position and velocity from geometric-feature and ego-motion histories, using the objective below.

The visual-training dataset $\mathcal D_V$ pairs histories ending at $t$ with latent targets $\mathbf z_t=E_\phi(\mathbf p_t)$ from the fixed privileged encoder, where $\mathbf p_t$ uses a fixed two-step (40\,ms) latency and no dropout, and auxiliary targets $\mathbf u_t\in\mathbb R^6$, consisting of standardized current camera-frame box position and velocity. Let $\widehat{\mathbf u}_t\in\mathbb R^6$ denote the auxiliary prediction. The temporal-training objective is
\begin{equation}
\begin{aligned}
\mathcal L_V&=\mathbb E_{\mathcal D_V}
\left[\omega_t\big(\ell_z(t)+\lambda_{\mathrm{aux}}\ell_{\mathrm{state}}(t)\big)\right],\\
\ell_z(t)&=\frac{1}{8}\sum_{j=1}^{8}\alpha_j
\left(\frac{\widehat z_{t,j}-z_{t,j}}{\sigma_j+\epsilon}\right)^2,\\
\ell_{\mathrm{state}}(t)&=\frac{1}{6}\|\widehat{\mathbf u}_t-\mathbf u_t\|_2^2.
\end{aligned}
\label{eq:visual_loss}
\end{equation}
Here $\sigma_j$ is the training-set standard deviation of latent coordinate $j$, and $\epsilon$ ensures numerical stability. The nonnegative weights $\alpha_j$, normalized to average one, reflect the fixed decoder's action sensitivity to one-standard-deviation latent errors. The coefficient $\lambda_{\mathrm{aux}}$ balances auxiliary supervision, and $\omega_t$ emphasizes close-range observations (Sec.~\ref{sec:training-settings}).

\paragraph{Visual-policy data aggregation}
Initial data come from privileged-controller rollouts. We then run the visual controller, label its visited states with the fixed privileged encoder, aggregate the data, and refit the GRU and output heads with the localizer and controller fixed. This second DAgger-style procedure~\cite{dagger} provides latent targets for the views encountered under visual control. Unlike the action-labeled aggregation in Stage~2, it supervises the predicted latent rather than the controller's action.

Cached geometric features, ego-motion, and privileged observations provide aligned sequences and latent labels. Windows stay within episodes; validation uses separate episodes, and aggregation-round selection and final evaluation use separate throws.

\paragraph{Appearance adaptation}
We augment box colors during localizer pretraining and rollout collection; localizer pretraining also applies random RGB gain and offset, pixel noise, small image shifts, and about 2.4\,cm depth noise, and the temporal model is fit on noise-perturbed features with the color-augmented localizer frozen. Motion blur, auto-exposure dynamics, and structured depth artifacts are not modeled.

\subsection{Downstream Carrying and Handover}
\label{sec:deployment}

The downstream demonstration in Fig.~\ref{fig:teaser} connects the catching policy to the pretrained SONIC motion controller~\cite{sonic} for walking with the retained box and handover. Visual human detection identifies the recipient for the handover stage.

A supervisor transfers command authority after persistent near-chest box evidence and low leg velocity, torso tilt, and base speed establish a settled hold. Hold-state persistence accommodates close-range occlusion. At transfer, the supervisor establishes the navigation frame and switches gains and observations to SONIC. The reference holds the current posture for 2\,s before transitioning to a standing cradle for walking and handover. A MuJoCo deployment twin checks perception, communication, joint commands, and controller transfer.

In the 20-trial downstream task evaluation, cumulative success rates are 65\% after catching, 35\% after carrying, 30\% after handover, and 25\% after return.

\section{Experiments}
\label{sec:experiments}

Our experiments address four questions:
\begin{itemize}
\item \textbf{Q1:} How well does \model catch in simulation and on the real robot, compared with privileged input and human catchers?
\item \textbf{Q2:} How do the catching objective and motion priors shape whole-body control, and how does catching vary with box size and payload?
\item \textbf{Q3:} How should the object-dynamics latent be learned, and how does it handle lateral offsets and perception delay?
\item \textbf{Q4:} Which visual inputs and supervision ground the latent in egocentric vision, and how do appearance, distance, and detector-based perception affect it?
\end{itemize}

In the tables, pink identifies proposed configurations, gray denotes references or auxiliary diagnostic rates, and bold marks the highest catch rate in each comparison, excluding gray references. The abbreviation w/o means without, and GT denotes ground truth.

\subsection{Experimental Setup}
\label{sec:training-settings}
\label{sec:protocol}
\label{sec:robustness}

\paragraph{Training settings}
Stage~1 uses randomized state observations; its teacher is queried with ground-truth box position and velocity, and the ground-truth reference feeds ideal box measurements through its Kalman filter. Stage~2 BC initialization uses five data-aggregation rounds and gives visible-box samples fivefold weight. Stage~2 joint PPO uses 4096 environments, 600 iterations, and an initial learning rate of $10^{-4}$, without additional latent noise or whole-vector dropout. Release distances span 2.5--5.0\,m; box size is randomized among three sizes: 25$\times$19$\times$13, 34$\times$26$\times$22, and 42$\times$32$\times$27\,cm.

The privileged encoder has two 64-unit residual hidden layers. The visual model combines eight geometric features and nine ego-motion features using a two-layer, 128-unit GRU. The color-augmented localizer is pretrained for 20 epochs with batch size 32 and learning rate $3\times10^{-4}$. Temporal training uses AdamW for 60 epochs with batch size 512 and maximum learning rate $2\times10^{-3}$. Sequence training uses $\lambda_{\mathrm{aux}}=0.5$ and an eightfold sample weight at box ranges of 0.5--2.5\,m, normalized within each minibatch. Visual training uses two or three DAgger rounds per controller, selected on held-out evaluation throws. On one RTX 3090, rendering-free Stage~2 PPO runs at about 28k environment steps/s (4096 environments), whereas rendering $192\times144$ RGB-D in the loop runs at about 1.5k steps/s (320 environments): the 59M steps of Stage~2 PPO take about 35\,min instead of roughly 11\,h, while Stage~3 renders about 1.6M steps.

\paragraph{Evaluation protocol and comparisons}
Each environment contributes one throw from 3.0--3.5\,m with a $34\times26\times22$\,cm box. Evaluation starts without a held box, scores from the first step, and ends after 640 steps before reset. Rethrows, deliberately unreachable targets, robot pushes, and held-box kicks are disabled. Reference start frames are uniformly sampled from the standing segment, independently of the policy.

Throw attitude, spin, aim scatter, physical properties, and reset-state noise remain randomized. A catch requires chest-region support, wrists flanking the box, clearance from head/shoulder support, and low box speed for 0.6\,s. Torso excursion records whether tilt reaches $38^{\circ}$ between launch and evaluation end, separately from catch success and falls.

The initialization comparison trains six controllers per latent configuration and evaluates each on 640 held-out throws (3840 throws per configuration); the controllers share the predecessor policy and, for our configuration, BC initialization. Tables report mean catch rates. The state-policy reference uses one checkpoint.

We compare the ground-truth state-policy reference and two latent configurations, each with privileged and visual inputs. Both inherit the decoder body and use the same architecture and control interface. Ours uses BC initialization; the state-initialized alternative aligns encoder coordinates with position/velocity and maps the predecessor's input weights without BC. Both continue joint encoder/decoder PPO training (Tab.~\ref{tab:initialization-ablation}). The privileged observation mask uses the hardware camera's field of view ($\pm53^{\circ}\times\pm40^{\circ}$), whereas the rendered 4:3 pinhole image spans $\pm45^{\circ}$ vertically, so privileged-input results are a reference for visual substitution. Tab.~\ref{tab:robustness}, Tab.~\ref{tab:latency}, and Fig.~\ref{fig:success-rates} use the first-trained of the six controllers (81.6--91.7\% at the reference): its default visual component in the figure (85.3\%) and its color-augmented component in the tables (88.4--88.8\%). In tables and figures, the state policy is the Stage~1 controller, privileged $\mathbf z$ is the Stage~2 latent controller with privileged box observations, and visual $\widehat{\mathbf z}$ (visual latent) is the full Stage~3 system, \model.

\paragraph{Condition benchmark}
Tab.~\ref{tab:robustness} varies one factor at a time from the setting of Sec.~\ref{sec:protocol}: a 0.18\,kg box thrown from 3.0--3.5\,m with at most $5^{\circ}$ aim error. Small and large sizes are measured real boxes; the regular size and the weight and distance tests use the $34\times26\times22$\,cm training box, and distance tests raise the launch-speed cap from at most 8.5\,m/s in training to about 12\,m/s, so 99\% of 5\,m and all 7\,m throws are faster than any training throw. The large box exceeds every training size, the small box is near the smallest (25$\times$19$\times$13\,cm), 0.5--2\,kg payloads and 7\,m throws lie outside the training conditions, and Stage~3 data use the 3.0--3.5\,m throw setting. Weight, distance, and color columns use the chest-region criterion of Sec.~\ref{sec:protocol}. Because box geometry changes, the size columns instead require the box to settle within 0.65\,m of the pelvis below 1.2\,m/s for 0.6\,s, stay above 0.35\,m, and remain at rest at 12.8\,s; early resets fail. Interception requires the box to slow below 1\,m/s near the torso, irrespective of retention. The visual-latent rows use the color-augmented visual component.

\subsection{Overall Performance}
\begin{table*}[t]
\centering
\caption{\textbf{Simulation robustness.} Catch and interception rates (\%); one checkpoint per controller, 640 throws per condition (320 for the detector). With the same detector, visual latent control exceeds detection plus filtering within its visual training range.}
\label{tab:robustness}
\papertablestyle
\setlength{\tabcolsep}{1.0pt}
\resizebox{\linewidth}{!}{% scale to the text width (long detector labels)
\begin{tabular}{@{}ll*{19}{c}@{}}
\toprule
 & & \multicolumn{4}{c}{Box size (cm)} & \multicolumn{4}{c}{Box weight (kg)} & \multicolumn{4}{c}{Throw distance (m)} & \multicolumn{7}{c}{Box color}\\
\cmidrule(lr){3-6}\cmidrule(lr){7-10}\cmidrule(lr){11-14}\cmidrule(lr){15-21}
Controller & Rate (\%) & \begin{tabular}[c]{@{}c@{}}Small\\{\fontsize{6.5}{7}\selectfont 22$\times$21$\times$15}\end{tabular} & \begin{tabular}[c]{@{}c@{}}Regular\\{\fontsize{6.5}{7}\selectfont 34$\times$26$\times$22}\end{tabular} & \begin{tabular}[c]{@{}c@{}}Large\\{\fontsize{6.5}{7}\selectfont 46$\times$41$\times$32}\end{tabular} & Avg. & 0.5 & 1.0 & 2.0 & Avg. & 2.5 & 5.0 & 7.0 & Avg. & White & Grey & Red & Green & Blue & Black & Avg. \\
\midrule
\multicolumn{2}{l}{\emph{Privileged box-state input}} & \multicolumn{19}{l}{}\\
Stage~1 / GT state & Catch & \tablenum{19.5} & \tablebestnum{89.4} & \tablenum{43.1} & \tablenum{50.7} & \tablenum{88.8} & \tablenum{85.2} & \tablenum{73.1} & \tablenum{82.4} & \tablenum{88.1} & \tablebestnum{79.7} & \tablebestnum{58.3} & \tablebestnum{75.4} & \tablebestnum{88.9} & \tablebestnum{88.9} & \tablebestnum{88.9} & \tablebestnum{88.9} & \tablebestnum{88.9} & \tablebestnum{88.9} & \tablebestnum{88.9} \\
 & \tableref{Intercept} & \tableref{\tablenum{54.2}} & \tableref{\tablenum{93.8}} & \tableref{\tablenum{79.4}} & \tableref{\tablenum{75.8}} & \tableref{\tablenum{95.5}} & \tableref{\tablenum{95.5}} & \tableref{\tablenum{93.0}} & \tableref{\tablenum{94.7}} & \tableref{\tablenum{93.3}} & \tableref{\tablenum{88.0}} & \tableref{\tablenum{65.3}} & \tableref{\tablenum{82.2}} & \tableref{\tablenum{93.8}} & \tableref{\tablenum{93.8}} & \tableref{\tablenum{93.8}} & \tableref{\tablenum{93.8}} & \tableref{\tablenum{93.8}} & \tableref{\tablenum{93.8}} & \tableref{\tablenum{93.8}} \\
\tableours{Stage~2 / privileged $\mathbf z$} & Catch & \tablebestnum{21.6} & \tablenum{88.4} & \tablebestnum{47.8} & \tablebestnum{52.6} & \tablebestnum{93.4} & \tablebestnum{92.5} & \tablebestnum{84.7} & \tablebestnum{90.2} & \tablebestnum{90.9} & \tablebestnum{79.7} & \tablenum{41.6} & \tablenum{70.7} & \tablenum{88.3} & \tablenum{88.3} & \tablenum{88.3} & \tablenum{88.3} & \tablenum{88.3} & \tablenum{88.3} & \tablenum{88.3} \\
 & \tableref{Intercept} & \tableref{\tablenum{46.6}} & \tableref{\tablenum{96.2}} & \tableref{\tablenum{86.7}} & \tableref{\tablenum{76.5}} & \tableref{\tablenum{97.0}} & \tableref{\tablenum{96.7}} & \tableref{\tablenum{93.9}} & \tableref{\tablenum{95.9}} & \tableref{\tablenum{97.0}} & \tableref{\tablenum{90.6}} & \tableref{\tablenum{49.5}} & \tableref{\tablenum{79.0}} & \tableref{\tablenum{96.2}} & \tableref{\tablenum{96.2}} & \tableref{\tablenum{96.2}} & \tableref{\tablenum{96.2}} & \tableref{\tablenum{96.2}} & \tableref{\tablenum{96.2}} & \tableref{\tablenum{96.2}} \\
\midrule
\multicolumn{2}{l}{\emph{Egocentric RGB-D input}} & \multicolumn{19}{l}{}\\
Detector (YOLO) + state-based & Catch & \tablenum{11.2} & \tablenum{68.1} & \tablenum{17.5} & \tablenum{32.3} & \tablenum{67.2} & \tablenum{59.7} & \tablenum{48.4} & \tablenum{58.4} & \tablenum{71.9} & \tablenum{60.0} & \tablenum{32.8} & \tablenum{54.9} & \tablenum{65.3} & \tablenum{10.9} & \tablenum{65.6} & \tablenum{62.5} & \tablenum{56.9} & \tablenum{41.6} & \tablenum{50.5}\\
 & \tableref{Intercept} & \tableref{\tablenum{45.6}} & \tableref{\tablenum{78.4}} & \tableref{\tablenum{70.9}} & \tableref{\tablenum{65.0}} & \tableref{\tablenum{80.9}} & \tableref{\tablenum{78.1}} & \tableref{\tablenum{72.8}} & \tableref{\tablenum{77.3}} & \tableref{\tablenum{80.9}} & \tableref{\tablenum{74.7}} & \tableref{\tablenum{41.2}} & \tableref{\tablenum{65.6}} & \tableref{\tablenum{77.2}} & \tableref{\tablenum{15.0}} & \tableref{\tablenum{80.9}} & \tableref{\tablenum{75.6}} & \tableref{\tablenum{68.1}} & \tableref{\tablenum{55.6}} & \tableref{\tablenum{62.1}} \\
Detector (our CNN) + state-based & Catch & \tablenum{20.5} & \tablenum{73.3} & \tablenum{24.4} & \tablenum{39.4} & \tablenum{75.6} & \tablenum{76.2} & \tablenum{65.5} & \tablenum{72.4} & \tablenum{85.0} & \tablenum{66.9} & \tablebestnum{38.3} & \tablenum{63.4} & \tablenum{77.3} & \tablenum{74.7} & \tablenum{74.5} & \tablenum{73.8} & \tablenum{77.2} & \tablenum{77.7} & \tablenum{75.9}\\
Detector (our CNN) + latent decoder & Catch & \tablenum{18.0} & \tablenum{85.2} & \tablenum{29.5} & \tablenum{44.2} & \tablenum{90.9} & \tablebestnum{91.2} & \tablebestnum{83.3} & \tablebestnum{88.5} & \tablenum{88.8} & \tablebestnum{67.7} & \tablenum{33.9} & \tablebestnum{63.5} & \tablenum{85.2} & \tablenum{87.0} & \tablebestnum{87.0} & \tablenum{86.4} & \tablebestnum{88.8} & \tablenum{87.0} & \tablenum{86.9}\\
\tableours{Stage~3 / visual $\widehat{\mathbf z}$ (ours)} & Catch & \tablebestnum{25.5} & \tablebestnum{88.8} & \tablebestnum{39.7} & \tablebestnum{51.3} & \tablebestnum{91.2} & \tablenum{89.7} & \tablenum{77.7} & \tablenum{86.2} & \tablebestnum{90.2} & \tablenum{43.0} & \tablenum{13.3} & \tablenum{48.8} & \tablebestnum{87.5} & \tablebestnum{87.7} & \tablenum{86.9} & \tablebestnum{87.2} & \tablenum{86.6} & \tablebestnum{87.8} & \tablebestnum{87.3}\\
 & \tableref{Intercept} & \tableref{\tablenum{58.1}} & \tableref{\tablenum{95.5}} & \tableref{\tablenum{86.1}} & \tableref{\tablenum{79.9}} & \tableref{\tablenum{97.2}} & \tableref{\tablenum{97.2}} & \tableref{\tablenum{91.9}} & \tableref{\tablenum{95.4}} & \tableref{\tablenum{97.3}} & \tableref{\tablenum{53.3}} & \tableref{\tablenum{18.6}} & \tableref{\tablenum{56.4}} & \tableref{\tablenum{95.5}} & \tableref{\tablenum{95.6}} & \tableref{\tablenum{94.2}} & \tableref{\tablenum{95.8}} & \tableref{\tablenum{95.5}} & \tableref{\tablenum{95.8}} & \tableref{\tablenum{95.4}} \\
\bottomrule
\end{tabular}}
\papertablenote{Avg.: unweighted mean within each condition group. Detector (YOLO): real-photo fine-tuned YOLOv8s on rendered $640\times480$ RGB-D; box depth feeds the state policy's Kalman filter. Simulation rows are single checkpoints, so differences below about 5 points are not resolved; the Regular column uses the $34\times26\times22$\,cm training box. Intercept: box slowed below 1\,m/s near the torso, without requiring retention. Binomial standard error: at most 2.0 points for 640 throws, 2.8 for 320. Detector (our CNN): our color-augmented heatmap CNN (same rendered training data and camera as ours) with a Kalman filter tuned on held-out throws and a ground-truth arm trigger; its state drives the state policy or, via $E_\phi$, our decoder.}
\end{table*}

\begin{table*}[t]
\centering
\caption{\textbf{Real-robot catching.} Catch rates (\%) over 20 G1 trials per condition, varying one factor at a time with the others fixed at about 2.5\,m and the regular box; 95\% Wilson intervals span up to about $\pm$20 points (e.g., 0.5 vs.\ 1\,kg: 9/20 vs.\ 12/20, Fisher $p=0.53$). The G1 reaches 76\% and 59\% of human success (three people, 60 trials per condition) at 2.5 and 5\,m.}
\label{tab:real}
\papertablestyle
\begin{tabular*}{\textwidth}{@{\extracolsep{\fill}}lcccccccccccc@{}}
\toprule
 & \multicolumn{4}{c}{Box size (cm)} & \multicolumn{4}{c}{Box weight (kg)} & \multicolumn{4}{c}{Throw distance (m)}\\
\cmidrule(lr){2-5}\cmidrule(lr){6-9}\cmidrule(lr){10-13}
Catcher & \begin{tabular}[c]{@{}c@{}}Small\\{\fontsize{6.5}{7}\selectfont 22$\times$21$\times$15}\end{tabular} & \begin{tabular}[c]{@{}c@{}}Regular\\{\fontsize{6.5}{7}\selectfont 31$\times$26$\times$21}\end{tabular} & \begin{tabular}[c]{@{}c@{}}Large\\{\fontsize{6.5}{7}\selectfont 46$\times$41$\times$32}\end{tabular} & Avg. & 0.5 & 1.0 & 2.0 & Avg. & 2.5 & 5.0 & 7.0 & Avg.\\
\midrule
\tableours{Real G1 (ours)} & \tablenum{20.0} & \tablenum{65.0} & \tablenum{45.0} & \tablenum{43.3} & \tablenum{45.0} & \tablenum{60.0} & \tablenum{30.0} & \tablenum{45.0} & \tablenum{65.0} & \tablenum{40.0} & \tablenum{15.0} & \tablenum{40.0}\\
\tableref{Human reference} & \tableref{\tablenum{75.0}} & \tableref{\tablenum{85.0}} & \tableref{\tablenum{78.3}} & \tableref{\tablenum{79.4}} & \tableref{\tablenum{85.0}} & \tableref{\tablenum{86.7}} & \tableref{\tablenum{76.7}} & \tableref{\tablenum{82.8}} & \tableref{\tablenum{85.0}} & \tableref{\tablenum{68.3}} & \tableref{\tablenum{28.3}} & \tableref{\tablenum{60.6}}\\
\bottomrule
\end{tabular*}
\end{table*}

For Q1 in simulation, the visual controller catches 3290/3840 throws (85.7\%), compared with 87.9\% using privileged latent input to the same decoder (Tab.~\ref{tab:initialization-ablation}), so visual prediction preserves most of the catch performance.

On the real robot, Tab.~\ref{tab:real} reports 20 trials per condition. Catch success reaches 65\%, 40\%, and 15\% for the regular box at 2.5, 5, and 7\,m and ranges from 30\% to 60\% across the tested payloads; with 20 trials, 95\% Wilson intervals span up to about $\pm$20 points. As a human reference, three people each caught 20 regular-box throws per distance, standing with arms relaxed until release. Pooled human success is 85.0\%, 68.3\%, and 28.3\% at 2.5, 5, and 7\,m; the real G1 reaches 76\% and 59\% of it at 2.5 and 5\,m (95\% intervals 54--107\% and 33--103\%), and both are low at 7\,m, which approaches the practical range of human throwing and catching.

\subsection{Whole-Body Control}
\label{sec:exp-wbc}

To answer Q2, we ablate the Stage~1 objective, reused in latent PPO, and examine box size and payload.

\paragraph{Reward objective}
\label{sec:reward-ablation}

\begin{table}[!htbp]
\centering
\caption{\textbf{Reward ablation.} Catch rates (\%, mean $\pm$ s.d. over three training seeds) after PPO. Removing all three task groups drops success to 33.7\% and 2.7\%. $^\ast$Foot-height termination; no episode reaches the throw. $^\ddagger$Stage~2 recipe trained by PPO from random initialization, one seed, 1000 iterations.}
\label{tab:ablation}
\papertablestyle
\begin{tabular*}{\linewidth}{@{\extracolsep{\fill}}p{0.49\linewidth}*{2}{>{\centering\arraybackslash}p{0.23\linewidth}}@{}}
\toprule
 & \multicolumn{2}{c}{Catch rate (\%), lateral offset (m)}\\
\cmidrule(lr){2-3}
Training objective & 0.0 & 0.6\\
\midrule
\tableref{Init.\ only (no PPO)} & \tableref{\tablenum{75.3}} & \tableref{\tablenum{44.2}}\\
\tableref{No Stage~1 (scratch)$^\ddagger$} & \tableref{\tablenum{6.7}} & \tableref{\tablenum{0.2}}\\
\tableours{Full objective} & \tablebestnum{88.6}{\fontsize{5.5}{6}\selectfont$\pm$4.8} & \tablebestnum{76.1}{\fontsize{5.5}{6}\selectfont$\pm$4.8}\\
\midrule
\multicolumn{3}{@{}l}{\emph{Task rewards removed}}\\
w/o success bonus & \tablenum{87.8}{\fontsize{5.5}{6}\selectfont$\pm$1.6} & \tablenum{71.8}{\fontsize{5.5}{6}\selectfont$\pm$2.6}\\
w/o approach shaping & \tablenum{82.8}{\fontsize{5.5}{6}\selectfont$\pm$5.0} & \tablenum{68.2}{\fontsize{5.5}{6}\selectfont$\pm$5.6}\\
w/o hold shaping & \tablenum{83.9}{\fontsize{5.5}{6}\selectfont$\pm$3.0} & \tablenum{50.8}{\fontsize{5.5}{6}\selectfont$\pm$3.8}\\
w/o all three task groups & \tablenum{33.7}{\fontsize{5.5}{6}\selectfont$\pm$5.0} & \tablenum{2.7}{\fontsize{5.5}{6}\selectfont$\pm$1.0}\\
\midrule
\multicolumn{3}{@{}l}{\emph{Motion priors removed}}\\
w/o tracking rewards & \tablenum{84.3}{\fontsize{5.5}{6}\selectfont$\pm$3.3} & \tablenum{71.5}{\fontsize{5.5}{6}\selectfont$\pm$0.2}\\
w/o AMP reward & \tablenum{83.8}{\fontsize{5.5}{6}\selectfont$\pm$2.0} & \tablenum{71.8}{\fontsize{5.5}{6}\selectfont$\pm$3.1}\\
w/o tracking and AMP & \multicolumn{2}{c}{training collapses$^\ast$}\\
\bottomrule
\end{tabular*}
\end{table}

Rewards are removed only during Stage~2 PPO from the shared initialization, so Tab.~\ref{tab:ablation} tests what retains and extends the skill. PPO raises success from 75.3\% to 88.6\% for centered throws and from 44.2\% to 76.1\% at 0.6\,m offset. Removing hold shaping reduces offset success to 50.8\%; removing all three task groups gives 2.7\%. Removing tracking or AMP alone costs 4--5 points, not significant over three seeds; removing both collapses training (a foot-height termination ends most episodes), and no episode reaches the throw. Without Stage~1, the same latent controller trained by PPO from random initialization with the Stage~2 recipe and a standard learning rate catches only 6.7\% and 0.2\% after 1000 iterations (6.7\% and 0.5\% after 600; one seed); Stage~1 itself was trained for longer, so this shows that it is needed within the Stage~2 budget.

\paragraph{Retention and torso motion}
Torso excursion ranges from 1\% to 72\% across the six visual controllers (mean 21\%, versus 14\% for the state-initialized group). One checkpoint catches 91.7\% but has 72\% excursion; another catches 86.2\% with 1\% excursion. These post-launch metrics distinguish successful box retention from a settled whole-body posture. For the six visual controllers, the pelvis never drops below 0.52\,m, and 3 of 3840 episodes end early (counted as failures).

\paragraph{Box size and payload}
With visual input, 0.5--2\,kg payloads are caught at 77.7--91.2\%, on par with the state policy (73.1--88.8\%; Tab.~\ref{tab:robustness}). Performance drops for boxes outside the reference size: visual catch success is 25.5\% for small boxes and 39.7\% for large boxes, compared with interception rates of 58.1\% and 86.1\%. The gap to interception indicates that many of these boxes are reached but not retained. The controller is effectively size-blind: Stage~1 trains on the regular box without a size input, $\mathbf z$ does not encode size, and small boxes fail with privileged input as well (19.5--21.6\%).

\subsection{Object Dynamics}
\label{sec:exp-dynamics}

To answer Q3, we compare latent initializations and test lateral offsets and added delay.

\paragraph{Controller initialization}
\label{sec:initialization-ablation}
The BC-initialized latent-learning pipeline improves visual catch success by 5.2 percentage points over position/velocity weight mapping (Tab.~\ref{tab:initialization-ablation}), with a similar gain under privileged input; the visual gain holds in five of six seed pairs. Action matching establishes the latent control interface before joint encoder/decoder PPO refinement. Ridge probes with quadratic features on held-out flight ticks decode the delayed box position and velocity from $\mathbf z$ ($R^2$ 0.94--0.99 and 0.90--0.99 across the six controllers) but not box size, orientation, or angular velocity ($R^2\le0.13$), whereas the weight-mapped latent keeps position and velocity linearly ($R^2\ge0.98$) and retains some orientation ($R^2$ up to 0.58). A probe of the box's current, undelayed state (first controller of each configuration) shows that $\mathbf z$ partly compensates for observation age: for samples at least 80\,ms old, it predicts the current position and velocity with RMSE 0.33\,m and 1.5\,m/s, against 0.6--0.7\,m and 3.4\,m/s for the delayed sample itself and 0.40\,m and 2.1\,m/s for the weight-mapped latent, which stays closer to the delayed observation.

\paragraph{Lateral landing offset}
\begin{figure}[t]
\centering
\includegraphics[width=\columnwidth]{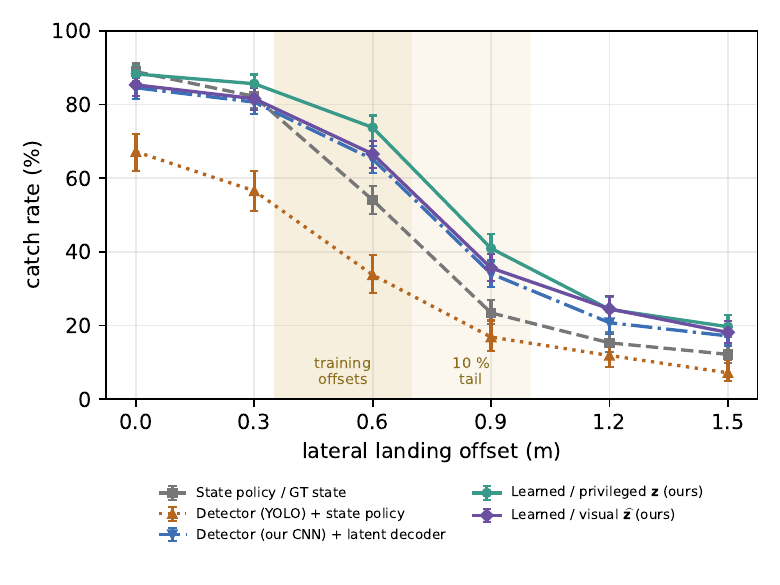}
\caption{\textbf{Catching across lateral landing offsets.} Catch rates for the ground-truth state-policy reference, the YOLO detector with the state policy, our CNN detector with the latent decoder, privileged latent input, and visual latent input. Shading indicates the training-offset region and its tail. Each point pools 640 throws for one policy checkpoint (320 throws for the YOLO curve). Visual latent input exceeds the ground-truth state policy by about 12 points at 0.6 and 0.9\,m, whereas YOLO detection lowers the state policy's catch rate at every reported offset.}
\label{fig:success-rates}
\vspace{-2pt}
\end{figure}

Figure~\ref{fig:success-rates} varies lateral offset for one checkpoint (640 throws per point). Visual success falls from 85.3\% at 0\,m to 66.6\% at 0.6\,m and 35.6\% at 0.9\,m; privileged-input rates are 88.3\%, 73.8\%, and 40.9\%. The state policy with ground-truth state falls faster, from 88.9\% to 54.1\% and 23.4\%, so the visual controller exceeds it by about 12 points at 0.6 and 0.9\,m (five further controllers: $+12.5\pm8.9$ points at 0.6\,m); with the YOLO detector the state policy reaches 67.2\%, 33.8\%, and 16.9\%. The margin over the state policy arises in Stage~2 PPO (44.2\% at 0.6\,m before PPO; Tab.~\ref{tab:ablation}), so Stage~1 is not an upper bound.

\paragraph{Added perception delay}
\begin{table}[!htbp]
\centering
\caption{\textbf{Latency.} Hardware timings (top) and delay sensitivity (bottom). Latency randomization helps at every delay.}
\label{tab:latency}
\papertablestyle
\begin{tabular*}{\linewidth}{@{\extracolsep{\fill}}lccc@{}}
\toprule
 & \multicolumn{3}{c}{Latency (ms)}\\
\cmidrule(lr){2-4}
Hardware measurement & p50 & p95 & p99\\
\midrule
Camera exposure $\rightarrow$ YOLO input & \tablenum{50.1} & \tablenum{69.0} & \tablenum{72.0}\\
YOLO inference & \tablenum{8.2} & \tablenum{12.9} & \tablenum{14.9}\\
Camera exposure $\rightarrow$ control tick & \tablenum{69.1} & \tablenum{90.2} & \tablenum{94.1}\\
Control period (nominal 20\,ms) & \tablenum{20.0} & \tablenum{20.4} & \tablenum{20.9}\\
\bottomrule
\end{tabular*}
\par\vspace{3pt}
\begin{tabular*}{\linewidth}{@{\extracolsep{\fill}}l*{7}{c}@{}}
\toprule
 & \multicolumn{7}{c}{Added perception delay (ms)}\\
\cmidrule(lr){2-8}
Catch rate (\%), simulation & 0 & 20 & 40 & 60 & 80 & 120 & 160\\
\midrule
\tableours{Stage~3 / visual $\widehat{\mathbf z}$ (ours)} & \tablenum{88.4} & \tablenum{87.2} & \tablenum{83.0} & \tablenum{72.7} & \tablenum{60.9} & \tablenum{32.3} & \tablenum{19.7}\\
w/o latency randomization & \tablenum{80.8} & \tablenum{79.1} & \tablenum{71.7} & \tablenum{65.8} & \tablenum{50.8} & \tablenum{27.0} & \tablenum{11.1}\\
\bottomrule
\end{tabular*}
\papertablenote{p50/p95/p99: 50th/95th/99th percentiles. Hardware: detector-based pipeline (YOLO, Kalman filter, Stage~1 policy), excluding actuation. Simulation delays encoder inputs beyond zero-latency rendering (latent targets already lag 40\,ms); each row: one training seed, 640 throws per point; reruns of a cell differ by up to 6 points.}
\end{table}

Tab.~\ref{tab:latency} delays the visual encoder input in simulation to test the Stage~2 latency randomization. With visual input, catch success is 88.4\%, 87.2\%, and 83.0\% with 0, 20, and 40\,ms of added delay and 72.7\% at 60\,ms, versus 65.8\% at 60\,ms for the controller trained without latency randomization (one training seed each), which is 5--11 points lower at every delay; beyond the training range, both decline (32.3\% and 27.0\% at 120\,ms). The 60 and 80\,ms columns bracket the 69\,ms median of our detector-based hardware pipeline, so the zero-delay rates elsewhere are likely optimistic for hardware. With privileged input at matched total delay (added delay plus the 40\,ms label lag), success falls from 88.3\% at 40\,ms to 76.4\% at 120\,ms, against 88.4\% to 60.9\% with visual input, so most of the visual decline arises in the encoder, whose training inputs are not delayed. The latency-randomized controller is most successful at 60\,ms of delay (91.9\% vs.\ 82.7\% without delay; privileged input), whereas the one trained without delay peaks at zero (84.5\%), consistent with the controller compensating for the latency it was trained with.

\begin{figure*}[t]
\centering
\includegraphics[width=\textwidth]{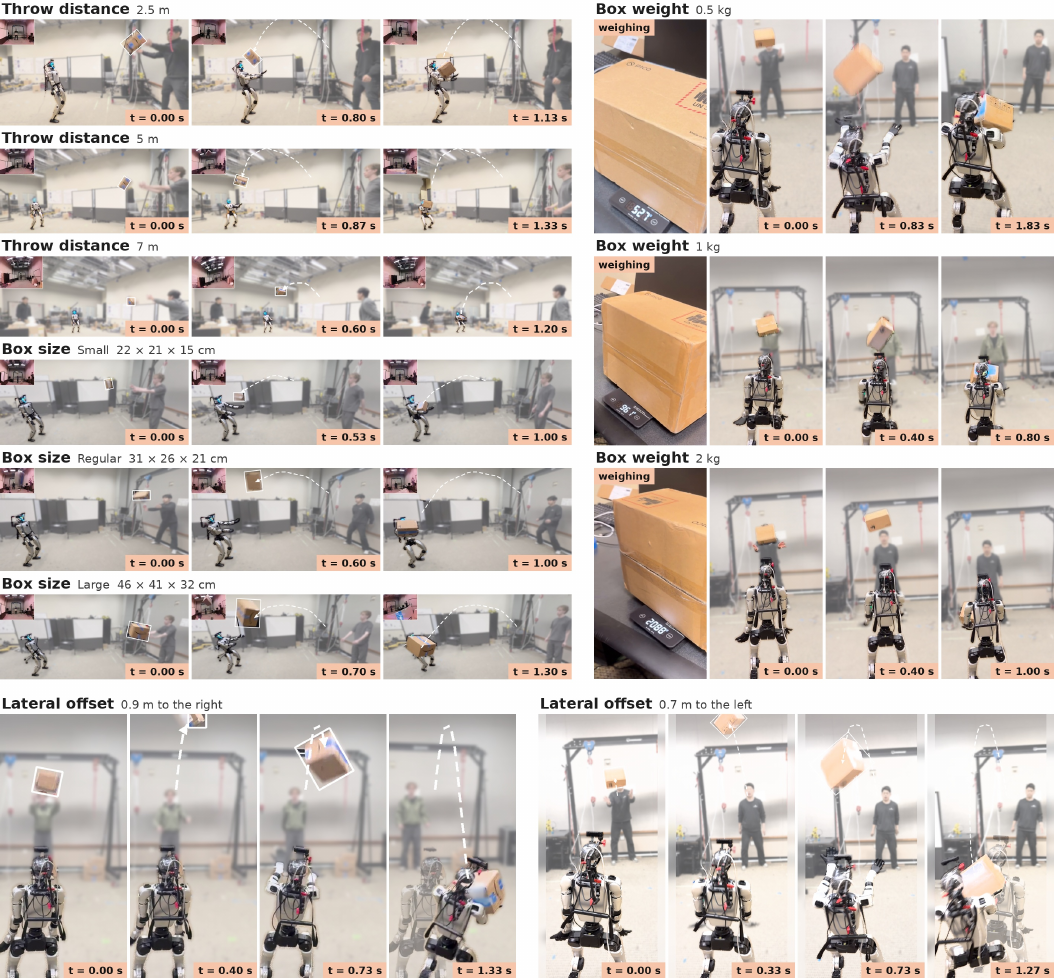}
\caption{\textbf{Real-robot catching across throw distances, box sizes, payloads, and lateral offsets.} Successful trials at 2.5, 5, and 7\,m and with small, regular, and large boxes (left, each at release, during approach, and after the catch), with 0.5, 1, and 2\,kg payloads (right; weighing views show the payload mass), and with lateral offsets of 0.9\,m right and 0.7\,m left (bottom). Insets show onboard camera views; timestamps indicate elapsed time within each sequence. White outlines mark box positions and dashed curves indicate flight direction. The same visual latent controller succeeds at every tested distance, box size, payload, and lateral offset; Tab.~\ref{tab:real} reports catch rates for distance, size, and payload, and Sec.~\ref{sec:real-robot} the lateral-offset rate.}
\label{fig:real-deployment}
\label{fig:real-weight-offset}
\vspace{-2pt}
\end{figure*}

\subsection{Visual Grounding}
\label{sec:exp-visual}

To answer Q4, we ablate the visual encoder and test appearance, distance, and a detector-based alternative.

\paragraph{Visual inputs and supervision}
\label{sec:visual-ablation}

\begin{table}[!htbp]
\centering
\caption{\textbf{Visual-encoder ablation.} Catch rates (\%, mean $\pm$ s.d. over five policies). Removing temporal history costs the most.}
\label{tab:visual_ablation}
\papertablestyle
\begin{tabular*}{\linewidth}{@{\extracolsep{\fill}}p{0.49\linewidth}*{2}{>{\centering\arraybackslash}p{0.23\linewidth}}@{}}
\toprule
 & \multicolumn{2}{c}{Catch rate (\%), lateral offset (m)}\\
\cmidrule(lr){2-3}
Visual encoder & 0.0 & 0.6\\
\midrule
\tableref{Privileged latent $\mathbf z$ (reference)} & \tableref{\tablenum{87.9}{\fontsize{5.5}{6}\selectfont$\pm$3.5}} & \tableref{\tablenum{74.3}{\fontsize{5.5}{6}\selectfont$\pm$4.5}}\\
\tableours{Full visual encoder} & \tablenum{85.2}{\fontsize{5.5}{6}\selectfont$\pm$4.2} & \tablenum{66.6}{\fontsize{5.5}{6}\selectfont$\pm$8.9}\\
\midrule
\multicolumn{3}{@{}l}{\emph{Inputs removed}}\\
w/o camera ego-motion & \tablenum{74.8}{\fontsize{5.5}{6}\selectfont$\pm$3.8} & \tablenum{44.0}{\fontsize{5.5}{6}\selectfont$\pm$3.2}\\
\shortstack[l]{w/o temporal history\\(single frame)} & \tablenum{49.6}{\fontsize{5.5}{6}\selectfont$\pm$4.9} & \tablenum{26.8}{\fontsize{5.5}{6}\selectfont$\pm$2.4}\\
\midrule
\multicolumn{3}{@{}l}{\emph{Training components removed}}\\
\shortstack[l]{w/o DAgger\\(teacher rollouts only)} & \tablebestnum{86.2}{\fontsize{5.5}{6}\selectfont$\pm$2.8} & \tablenum{67.6}{\fontsize{5.5}{6}\selectfont$\pm$4.4}\\
w/o box-state loss & \tablenum{85.0}{\fontsize{5.5}{6}\selectfont$\pm$3.8} & \tablebestnum{70.3}{\fontsize{5.5}{6}\selectfont$\pm$4.1}\\
w/o heatmap supervision & \tablenum{77.8}{\fontsize{5.5}{6}\selectfont$\pm$5.5} & \tablenum{55.7}{\fontsize{5.5}{6}\selectfont$\pm$7.0}\\
\shortstack[l]{w/o latent regression\\(state regression instead)} & \tablenum{66.5}{\fontsize{5.5}{6}\selectfont$\pm$7.1} & \tablenum{47.1}{\fontsize{5.5}{6}\selectfont$\pm$2.8}\\
\bottomrule
\end{tabular*}
\end{table}

Tab.~\ref{tab:visual_ablation} evaluates visual inputs and training objectives over five trained policies per variant. Variants use one aggregation round, except teacher-rollout training without DAgger. A single-frame encoder, which cannot observe motion, loses 35.6 and 39.8 percentage points on the centered and 0.6\,m cells, and regressing the 26-D privileged state for the frozen encoder instead of the latent loses 18.7 and 19.5. Removing camera ego-motion costs 10.4 and 22.6 points, and removing heatmap supervision 7.4 and 10.9. Teacher-rollout training alone achieves 86.2\% and 67.6\% at the two offsets, while removing the auxiliary box-state loss gives 85.0\% and 70.3\%. Both stay within seed variation, and zero or one aggregation round gives 86.2 and 85.2\% on the centered cell, against 85.7\% for the main encoders (two or three rounds) of the same five seeds, so neither component measurably helps in simulation. The state-regression variant must also predict orientation, angular velocity, and extents that $\mathbf g_t$ barely observes, so its gap does not isolate the latent target.

\paragraph{Box appearance}

\begin{table}[!htbp]
\centering
\caption{\textbf{Color augmentation.} Catch rates (\%), decoder fixed. Augmentation raises blue and black boxes to 88.0 and 88.8\%.}
\label{tab:color-ablation}
\papertablestyle
\begin{tabular*}{\linewidth}{@{\extracolsep{\fill}}p{0.49\linewidth}*{2}{>{\centering\arraybackslash}p{0.23\linewidth}}@{}}
\toprule
 & \multicolumn{2}{c}{Catch rate (\%)}\\
\cmidrule(lr){2-3}
Box appearance & \shortstack{Without color\\augmentation} & \tableours{\shortstack{With color\\augmentation}}\\
\midrule
Cardboard & \tablenum{87.0} & \tablebestnum{87.8}\\
White & \tablenum{78.0} & \tablebestnum{87.0}\\
Grey & \tablenum{78.8} & \tablebestnum{88.9}\\
Red & \tablenum{87.3} & \tablebestnum{88.8}\\
Green & \tablenum{83.3} & \tablebestnum{87.8}\\
Blue & \tablenum{60.2} & \tablebestnum{88.0}\\
Black & \tablenum{61.2} & \tablebestnum{88.8}\\
\bottomrule
\end{tabular*}
\end{table}

\begin{table}[!htbp]
\centering
\caption{\textbf{Stage~2 initialization.} Catch rates (\%, mean $\pm$ s.d. over six controllers). Action matching averages 5 points higher.}
\label{tab:initialization-ablation}
\papertablestyle
\begin{tabular*}{\linewidth}{@{\extracolsep{\fill}}p{0.40\linewidth}*{2}{>{\centering\arraybackslash}p{0.27\linewidth}}@{}}
\toprule
 & \multicolumn{2}{c}{Catch rate (\%), latent input}\\
\cmidrule(lr){2-3}
Latent initialization & Stage~2 / privileged $\mathbf z$ & Stage~3 / visual $\widehat{\mathbf z}$\\
\midrule
\tableours{Action matching (ours)} & \tablebestnum{87.9}{\fontsize{5.5}{6}\selectfont$\pm$3.2} & \tablebestnum{85.7}{\fontsize{5.5}{6}\selectfont$\pm$3.4}\\
Weight mapping & \tablenum{82.8}{\fontsize{5.5}{6}\selectfont$\pm$3.2} & \tablenum{80.5}{\fontsize{5.5}{6}\selectfont$\pm$2.9}\\
\bottomrule
\end{tabular*}
\end{table}

With the decoder fixed, color augmentation raises success on blue/black boxes from 60.2/61.2\% to 88.0/88.8\%, while cardboard changes from 87.0\% to 87.8\% (Tab.~\ref{tab:color-ablation}). Across seven tested colors, success is 87.0--88.9\% in this evaluation run (86.6--87.8\% on six colors in the separate run of Tab.~\ref{tab:robustness}), demonstrating reduced appearance sensitivity.

\paragraph{Learned latent versus detection plus filtering}
The visual gap grows with throw distance: at 5\,m, privileged latent input matches the state policy (79.7\%), whereas visual input reaches 43.0\%, although the encoder still detects the box on every 5 and 7\,m throw. For a like-for-like comparison with detection plus filtering, two baselines use our own color-augmented heatmap CNN (same rendered training data, camera, and resolution) and pass its per-frame 3D box point to a Kalman filter with gravity and drag whose speed bound, noise, gating, and output lag are selected on held-out throws; both receive a ground-truth arm trigger, which raises the arms earlier than ours. The filtered state drives either the state policy (``Detector (our CNN) + state-based'' in Tab.~\ref{tab:robustness}) or, encoded by $E_\phi$, our decoder (``Detector (our CNN) + latent decoder''), which isolates the learned temporal model. Excluding the 5 and 7\,m throws, which lie beyond the Stage~3 data, the visual latent outperforms the state-based pipeline in all 13 conditions of Tab.~\ref{tab:robustness} (by 5.0--15.6 points) and the latent-decoder pipeline in 14 of 18 conditions of Tab.~\ref{tab:robustness} and Fig.~\ref{fig:success-rates} (mean +1.5 points, largest for box size; with the same trigger, 88.3 vs.\ 84.5\% at the reference and 69.8 vs.\ 65.2\% at 0.6\,m; across five further controllers at 0 and 0.6\,m the margin is $+0.3\pm5.2$ points, ahead in 7 of 10 cells), so its advantage over detection plus filtering comes from the Stage~2 latent interface, while the learned filter is on par with a tuned Kalman filter. At 5 and 7\,m the filtered pipelines catch more (67.7 and 33.9\% for the latent decoder), but this reflects the Stage~3 data range: adding 2.5--7\,m throws to Stage~3 raises visual success to 75.8 and 36.7\%, with 88.8\% at the reference. The ``Detector (YOLO)'' row reproduces our previously deployed perception node (real-photo weights, hardware depth and size checks, untuned filter) and serves only as a reference.

\FloatBarrier
\subsection{Real-Robot Evaluation}
\label{sec:real-robot}

To test the conditions of Q2--Q4 on hardware, we evaluate catches of human-thrown boxes with different sizes, payloads, throw distances, and lateral offsets. Unless the size is varied, real-robot trials use the $31\times26\times21$\,cm regular box.

Figure~\ref{fig:real-deployment} shows qualitative catches across these conditions. Human throwers cannot place the box at a precise lateral offset; of 20 throws aimed to either side (e.g., 0.9\,m right and 0.7\,m left in Fig.~\ref{fig:real-deployment}), the G1 caught 8.

\paragraph{Hardware runtime} Tab.~\ref{tab:latency} reports measurements from two hardware sessions of our detector-based pipeline (1274 box measurements and approximately 33,000 control ticks). The 95th-percentile camera-to-control latency is 90.2\,ms, and the 50\,Hz control period exceeds 25\,ms on 0.015\% of ticks.

%\FloatBarrier
\section{Discussion and Limitations}
\label{sec:discussion}

The decomposition lets visual grounding be learned with the controller fixed and keeps rendering out of reinforcement learning, so only about 1.6M rendered steps are needed. During latent PPO, retaining the catching skill requires the task rewards, chiefly hold shaping, and at least one motion prior; payloads up to 2\,kg remain catchable, whereas boxes outside the reference size are often intercepted but not retained, and torso excursion separates retention from a settled posture. For object dynamics, BC initialization improves visual catching by 5.2 points on average, and in a single-seed comparison, latency randomization adds 5--11 points across simulated delays. For visual grounding, temporal history and ego-motion contribute most, regressing the learned latent gives higher success than the evaluated full-state regression baseline, and, with the same detector, the learned latent outperforms detection plus Kalman filtering within its visual training range and, once Stage~3 data cover them, also on 5 and 7\,m throws. The main limitation is long-range visual grounding: on 5\,m throws, faster than any training throw and beyond the 3.0--3.5\,m Stage~3 data, visual success falls to 43.0\% while privileged input reaches 79.7\%, and real-robot success falls to 40\% and 15\% at 5 and 7\,m; adding far throws to the Stage~3 data raises simulated 5\,m success to 75.8\%, not yet tested on hardware. Further limitations: a single task with human throws and 20 real trials per condition (intervals up to about $\pm$20 points); hardware latency was measured only on our detector-based pipeline, where it exceeds the randomized range; the controller is effectively size-blind; and rendering omits motion blur and exposure dynamics.

\section{Conclusion}

\model learns visual reactive whole-body catching on a humanoid by decoupling whole-body control, object dynamics, and visual grounding into three stages linked by an object-dynamics latent. In simulation, visual control reaches 85.7\% catch success versus 87.9\% with privileged input, and on a real Unitree G1 it catches human-thrown boxes at 76\% and 59\% of human success at 2.5 and 5\,m. Extending this decomposition to other visual reactive whole-body tasks is a natural next step.

% arXiv version: references start on a new page.
\ifdefined\CatchManArxivVersion\clearpage\fi
\bibliographystyle{IEEEtran}
\bibliography{references}

\end{document}